\def\year{2020}\relax
\documentclass[letterpaper]{article} 
\usepackage{aaai20}  
\usepackage{times}  
\usepackage{helvet} 
\usepackage{courier}  
\usepackage[hyphens]{url}  
\usepackage{graphicx} 
\usepackage{graphicx}  
\usepackage{eucal}
\usepackage{amsmath}
\usepackage{enumerate}
\usepackage{tabularx}
\usepackage{hhline}
\usepackage{subcaption}
\usepackage{multirow}
\usepackage{booktabs} 
\usepackage{relsize}
\usepackage{enumitem}

\newcommand{\model}{{AFN}~}
\newcommand{\modelns}{{AFN}}
\newsavebox{\tempbox}

\title{Adaptive Factorization Network: Learning Adaptive-Order Feature Interactions}

\author{Weiyu Cheng, Yanyan Shen\thanks{Corresponding Author}, Linpeng Huang\\
Shanghai Jiao Tong University
\\
\{weiyu\_cheng, shenyy, lphuang\}@sjtu.edu.cn}

\begin{document}

\maketitle

\begin{abstract}

Various factorization-based methods have been proposed to leverage second-order, or higher-order cross features for boosting the performance of predictive models. They generally enumerate all the cross features under a predefined maximum order, and then identify useful feature interactions through model training, which suffer from two drawbacks. First, they have to make a trade-off between the expressiveness of higher-order cross features and the computational cost, resulting in suboptimal predictions. Second, enumerating all the cross features, including irrelevant ones, may introduce noisy feature combinations that degrade model performance. In this work, we propose the Adaptive Factorization Network (AFN), a new model that learns arbitrary-order cross features adaptively from data. The core of AFN is a logarithmic transformation layer that converts the power of each feature in a feature combination into the coefficient to be learned. The experimental results on four real datasets demonstrate the superior predictive performance of AFN against the state-of-the-arts.
\end{abstract}

\section{Introduction}

Feature engineering is typically recognized as central to successful machine learning tasks, such as recommender systems~\cite{Lian:2017:PLJ:3124791.3124794}, computational advertising~\cite{DBLP:conf/kdd/HePJXLXSAHBC14} and search ranking~\cite{DBLP:journals/corr/LianX16}. 
Except for exploiting raw features, it is usually crucial to find effective transformations of raw features to boost the performance of predictive models.
\emph{Cross features} are a major type of feature transformations, where multiplication is performed over sparse raw features to form new features~\cite{wide_and_deep}.
However, handcrafting useful cross features is inevitably expensive and time-consuming, and the results may not generalize to unseen feature interactions.
In order to solve this problem, Factorization Machines (FMs)~\cite{FM,libFM} are proposed to explicitly model second-order cross features by parameterizing the weight of a cross feature as the inner product of the embedding vectors of the raw features. To be more general, higher-order FMs (HOFMs) involving higher-order feature combinations were also introduced in the original work~\cite{FM}.

Despite the superior predictive power, there remain two critical questions in FMs/HOFMs to be answered. First, \emph{what is the maximum order of cross features we should consider}? While a larger order enables the modeling of more complex feature interactions and seems to be beneficial, the number of cross features can increase exponentially with the value of the maximum order, resulting in high computational complexity.
Some recent works~\cite{hofm} focus on reducing the time complexity of training HOFMs. However, due to the large model size, the time cost of model training and prediction is still high for a large maximum order, which limits the practical usage of higher-order cross features.

Second, \emph{what is the set of useful cross features under the maximum order}? It is important to recognize that not all the features contain useful signals for estimating the target, and different cross features usually have different predictive power. Interactions among irrelevant features can be considered as noises, which have no contribution to the prediction or even degrade model performance.
To deal with this problem, Xiao et al. (\citeyear{AFM}) proposed Attentional Factorization Machines (AFM) to distinguish the importance of factorized interactions by reweighing each cross feature with an attention score~\cite{DBLP:journals/corr/BahdanauCB14}. The influence of useless cross features can be compromised by assigning lower weights. Nevertheless, applying the attention mechanism over complex feature combinations increases the computational cost significantly. As such, AFM aims at modeling second-order feature interactions only.

In this paper, we argue that existing factorization methods fail to answer the above two questions appropriately. In general, they follow an \emph{enumerating-and-filtering} manner to model feature interactions for prediction. The typical procedure is to predefine the maximum order, enumerate all the cross features within the maximum order, and then filter irrelevant cross features via training. This procedure consists of two major drawbacks. 
First, predefining a maximum order (which is typically small) restricts model's potential in finding discriminative cross features, because of the trade-off between expressive higher-order cross features and computational complexity.
Second, considering all the cross features may introduce noises and degrade the prediction performance, since not all the useless cross features can later be filtered out successfully.

To this end, we propose the Adaptive Factorization Network (\modelns) to learn arbitrary-order cross features and their weights adaptively from data. The key idea is to encode feature embeddings into a logarithmic space and convert the powers of features into the multiplications with coefficients. 
The core of \model is a logarithmic neural transformation layer consisting of multiple vector-wise logarithmic neurons. The purpose of each logarithmic neuron is to automatically learn the powers (i.e., orders) of the features in a possibly useful combination. 
Upon the logarithmic neural transformation layer, we apply the feedforward neural network to model element-wise feature interactions.
Different from FMs/HOFMs, \model is able to learn useful cross features from data adaptively, and the maximum order can be delivered on the fly. 
We summarize the major contributions of this paper as follows. 
\begin{itemize}
\item To the best of our knowledge, we are the first to introduce the logarithmic transformation structure with neural networks to model arbitrary-order feature interactions for prediction.
\item Based on the proposed logarithmic transformation layer, we propose the Adaptive Factorization Network (\modelns) to learn arbitrary-order cross features and their weights adaptively from data.
\item We show that FMs/HOFMs can be interpreted as two specializations of AFN, and the learned orders in AFN allow rescaling feature embeddings in different cross features.
\item We conducted extensive experiments on four public datasets. The results demonstrate that the orders of the learned cross features span across a wide range, and our approach achieves superior prediction performance compared with the state-of-the-art methods.
\end{itemize}

\section{Background}
\label{sec:background}

\subsubsection{Feature Embeddings.}
In many real-world predictive tasks such as CTR prediction, input instances consist of both sparse categorical features and numerical features.
By tradition, we represent each input instance as a sparse vector:
\begin{equation}
	\mathbf{x} = [\mathbf{x}_1,\mathbf{x}_2,...,\mathbf{x}_m]
\end{equation}
where $m$ is the number of \emph{feature fields} (e.g., item brand or user age), $\mathbf{x}_i$ is the representation of the $i$-th feature field (aka a \emph{feature}), and $\mathbf{x}$ is the concatenation of $\mathbf{x}_i$.
Since most categorical features are sparse and high-dimensional, a common practice is to map them into dense vectors (i.e., embeddings) in low-dimensional latent space. 
Specifically, a categorical feature $\mathbf{x}_i$ is initially a one-hot encoded vector, and we have its embedding $\mathbf{e}_i$ computed as follows:
\begin{equation}
	\mathbf{e}_i = \mathbf{V}_i\mathbf{x}_i
\end{equation}
where $\mathbf{V}_i$ denotes the embedding matrix of field $i$. For a numerical feature ${\bf x}_j$, its representation is a scalar $x_j$. To capture the interactions between numerical and categorical features, $x_j$ is also transformed into a dense vector in the same low-dimensional space:
\begin{equation} 
	\mathbf{e}_j = \mathbf{v}_jx_j
\end{equation}
where $\mathbf{v}_j$ is the embedding vector for the numerical field $j$. The resultant collection of feature embeddings $\mathbf{e}=\{\mathbf{e}_1,\mathbf{e}_2,...,\mathbf{e}_m\}$ will be used in FMs or neural networks for prediction~\cite{DBLP:conf/ijcai/ChengSZH18,DBLP:conf/icdm/LiSZ18}.

\subsubsection{Factorization Machines.}
Factorization Machines~\cite{FM} (FMs) are proposed to explicitly model second-order feature interactions for high-dimensional data.
Formally, the prediction of FMs is made as follows:
\begin{equation}
y = \langle \mathbf{w},\mathbf{x} \rangle+\sum_{j_2>j_1}^m \langle \mathbf{e}_{j_1},\mathbf{e}_{j_2} \rangle
\end{equation}
where $\langle\cdot,\cdot\rangle$ denotes the inner product operation. Intuitively, the first term $\langle \mathbf{w},\mathbf{x} \rangle$ is the linear combination of raw features, and the second term is the sum of pair-wise inner products of feature embeddings.
Higher-Order Factorization Machines (HOFMs) were introduced to capture higher-order feature interactions for prediction:
\small
\begin{equation}\label{eq:hofm}
\begin{aligned}
	y = \langle \mathbf{w},\mathbf{x} \rangle+\sum_{j_2>j_1}^m \langle \mathbf{e}_{j_1}^{(2)},\mathbf{e}_{j_2}^{(2)} \rangle+
	\sum_{j_3>j_2>j_1}^m \langle \mathbf{e}_{j_1}^{(3)},\mathbf{e}_{j_2}^{(3)},\mathbf{e}_{j_3}^{(3)} \rangle+\\
	\cdots+\sum_{j_n>\cdots>j_1}^m \langle \mathbf{e}_{j_1}^{(n)},\ldots,\mathbf{e}_{j_n}^{(n)} \rangle
\end{aligned}
\end{equation}
\normalsize
where the inner product operation is extended to represent the sum of element-wise products of multiple feature embeddings, and $n$ is the maximum order of cross features. 
Computing Equation (\ref{eq:hofm}) directly takes ${\rm O}(km^n)$ time, where $k$ is the rank of feature embeddings. Due to the high computational complexity, HOFMs have seldom been applied to real predictive systems~\cite{hofm}.

A common limitation of FMs and HOFMs is that they model all feature interactions with the same weight. As not all the cross features are useful, incorporating all of them for prediction may introduce noises and degrade model performance. As described above, some efforts have been devoted to alleviating this problem by exploiting attention mechanisms to assign non-uniform weights on different cross features~\cite{AFM}, or by learning weights for only retained cross features~\cite{xDeepFM}. However, these methods introduce additional costs and are still limited to a preset maximum order $n$ of feature interactions before model training. In practice, $n$ is usually set to a small value to make model size moderate. Such a design hinders the opportunity of finding discriminative higher-order cross features.
In this paper, we propose to learn arbitrary-order cross features adaptively from data. Both the maximum order and the set of cross features used for prediction will be identified adaptively through model training, leading to high computational efficiency without sacrificing predictive power.

\subsubsection{Logarithmic Neural Network (LNN).}
%
\begin{figure}[t]
	\centering  
	\includegraphics[width=0.65\columnwidth]{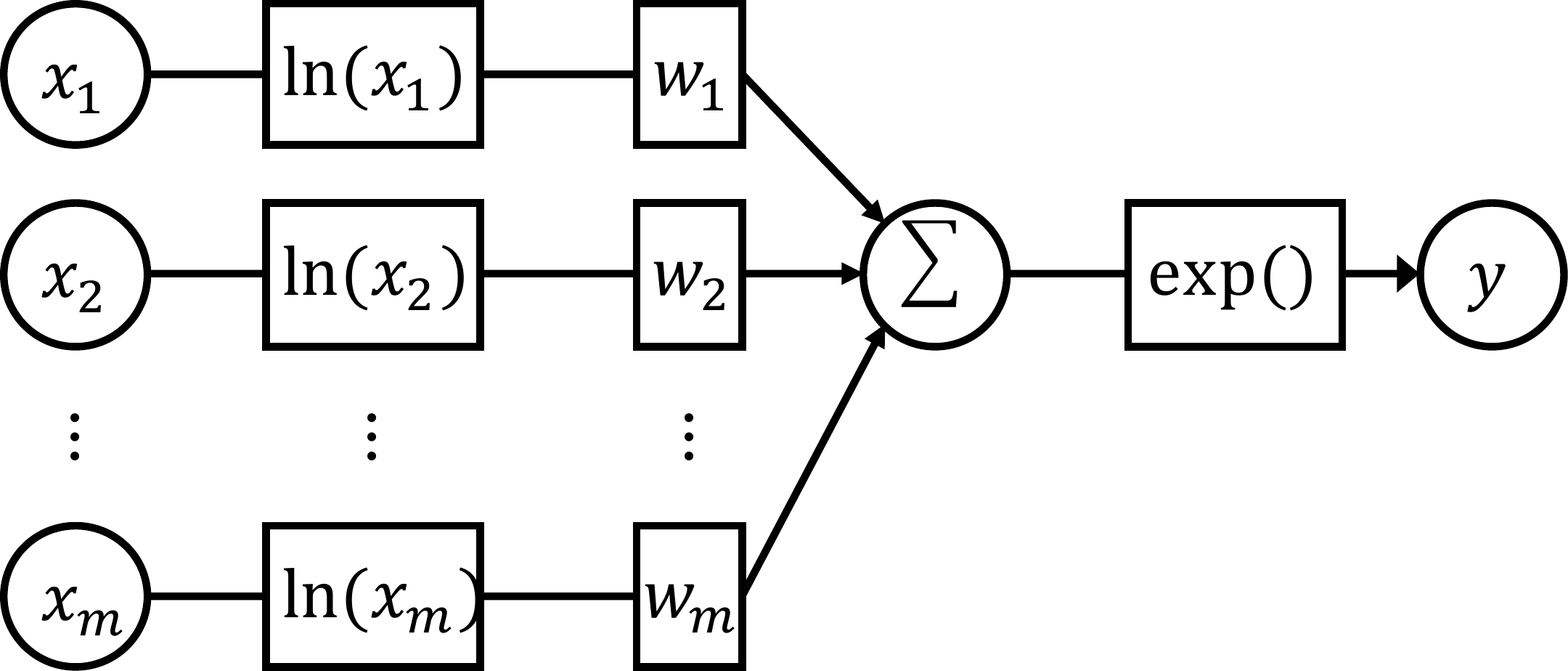} 
	\caption{Logarithmic neuron in LNN.}
	\label{fig:log_neuron}
\end{figure}
Logarithmic Neural Network~\cite{lnn} was initially proposed to approximate unbounded non-linear functions. LNN is composed of multiple logarithmic neurons, the structure of which is shown in Figure~\ref{fig:log_neuron}.
Formally, a logarithmic neuron can be formulated as:
\begin{equation}
\label{eq:lnn}
y = \exp(\sum_i w_i\ln x_i)
=\prod_i x_i^{w_i}
\end{equation}
The idea of LNN is to transform input into the logarithmic space, which converts multiplication to addition, division to subtraction, and powers to multiplication by a constant.
Although multi-layer perceptrons (MLPs) are known to be universal approximators, they have limited ability in approximating some functions such as multiplication, division, and powers when the input is unbounded~\cite{lnn}.
On the contrary, LNN is able to approximate such functions over the entire input range well.

In this paper, we exploit logarithmic neurons  
to adaptively learn the powers of each field in cross features from data.
We highlight three key differences between LNN and our proposed \modelns. 1) The learned power in \model is applied at a vector-wise level and shared among all feature embeddings in the same field. 2) The inputs to our model are feature embeddings to be learned. Thus we need to use some techniques to keep gradients stable and learn appropriate feature embeddings and combinations for prediction.
3) In \modelns, we further apply feed-forward hidden layers upon the learned cross features to enhance the expressiveness of our model.

\section{Adaptive Factorization Network}
 
\begin{figure}[t]
 	\centering 
 	\includegraphics[width=\columnwidth]{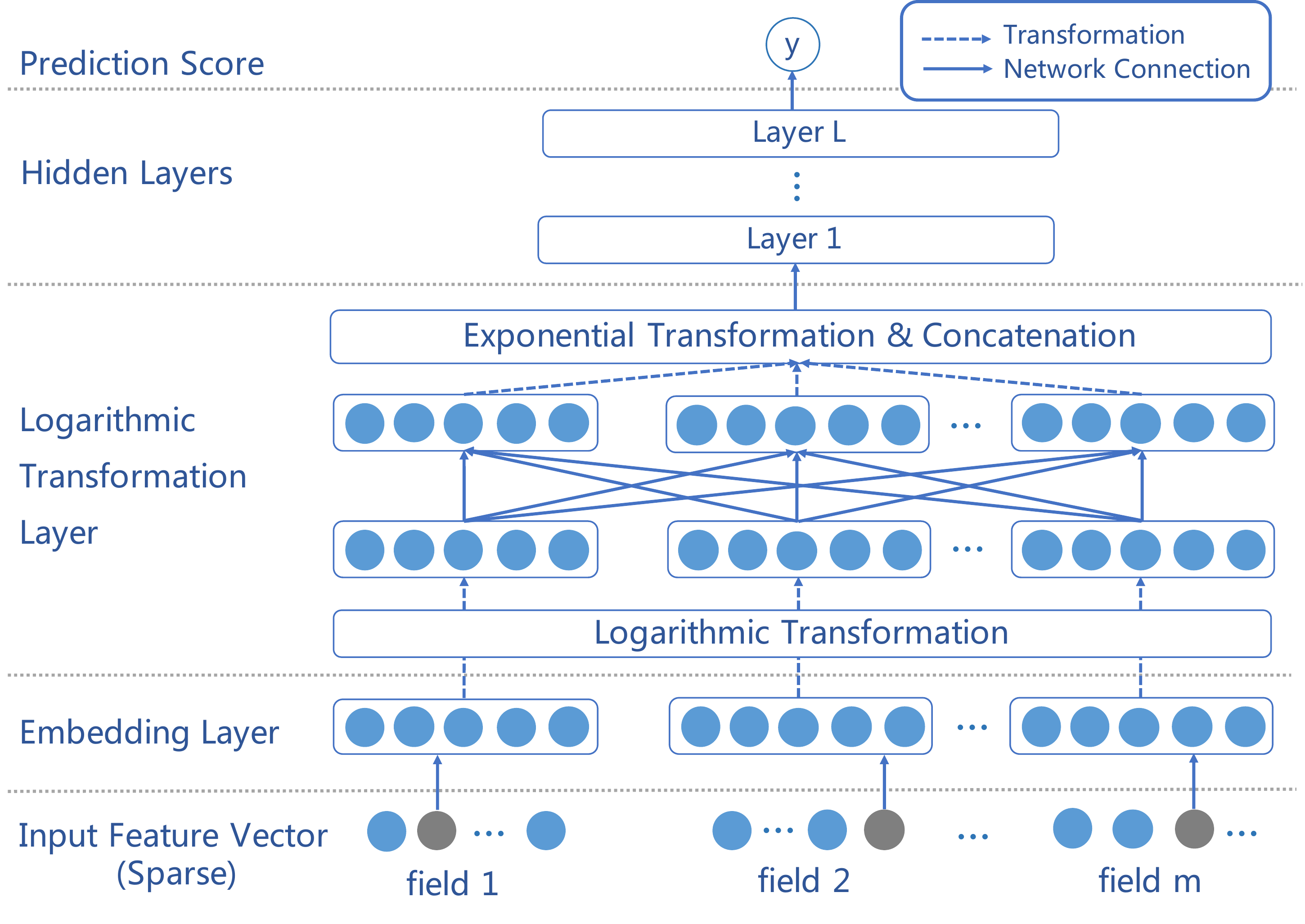}
 	\caption{Framework of \modelns.}
 	\label{fig:framework}
 \end{figure}

We first elaborate the \model model that learns adaptive-order feature interactions, including the optimization procedure and its ensemble with deep neural networks. We then make discussions on the learned feature orders in \modelns, the model's relation to FMs/HOFMs, and the time complexity. The overall structure of \model is depicted in Figure~\ref{fig:framework}.

\subsection{Model Architecture}
    
\subsubsection{Input Layer and Embedding Layer.}

The input layer of \model absorbs both sparse categorical features and numerical features. 
As described in Section~\ref{sec:background}, all the raw input features are first transformed into embeddings in a shared latent space. Here we introduce two crucial techniques for implementing the embedding layer. 
First, since we will apply a logarithmic transformation to feature embeddings in the successive layer, we need to keep all the values in the embeddings to be positive. Second, it is advised to add a small positive value $\epsilon$ (e.g., 1e-7) to zero embeddings to avoid numerical overflow.
After that, the output of the embedding layer is a collection of positive feature embeddings $\mathbf{e}=\{\mathbf{e}_1,\mathbf{e}_2,...,\mathbf{e}_m\}$.
 
\subsubsection{Logarithmic Transformation Layer}

The core of \model is the logarithmic transformation layer which learns the powers (i.e., orders) of each feature field in cross features.
This layer consists of multiple vector-wise \emph{logarithmic neurons}. Similar to Equation~(\ref{eq:lnn}), the output of the $j$-th vector-wise logarithmic neuron can be formulated as:
\begin{equation}
\label{eq:LTL}
\mathbf{y}_j = \exp(\sum_{i=1}^m w_{ij}\ln \mathbf{e}_i )
=\mathbf{e}_1^{w_{1j}}\odot\mathbf{e}_2^{w_{2j}}\odot...\odot\mathbf{e}_m^{w_{mj}}
\end{equation}
where $w_{ij}$ is the coefficient of the $j$-th neuron on the $i$-th field. The functions $\ln(\cdot)$ and $\exp(\cdot)$ and the power term $w_{ij}$ are all applied at the element-wise level to the corresponding vectors, and $\odot$ denotes element-wise product operation.
The main observation based on Equation~(\ref{eq:LTL}) is that the output of each logarithmic neuron $\mathbf{y}_j$ is able to represent any cross features. For example, when ${w_{1j}}$ and ${w_{2j}}$ are set to $1$, and ${w_{ij}} \ (2<i<m)$ are set to $0$, we have $\mathbf{y}_j=\mathbf{e_1}\odot\mathbf{e}_2$, which is a second-order cross feature for the first two raw feature fields.
Thus we can use multiple logarithmic neurons to obtain different feature combinations in arbitrary orders as the output of this layer.
Note that the elements in the coefficient matrix $\mathbf{W}_{LTL}\in \mathbf{R}^{m\times N}$ (where $N$ denotes the number of logarithmic neurons in the layer) are learnable parameters, and are not necessarily converged to be $0$s or $1$s.  

\subsubsection{Feed-forward Hidden Layers and Prediction.}

Upon the logarithmic transformation layer, we stack several fully-connected hidden layers to combine the formed cross features. We first concatenate all the cross features as input to the feed-forward network:
\begin{equation}
		\mathbf{z}_0 = [\mathbf{y}_1,\mathbf{y}_2,...,\mathbf{y}_N] 
\end{equation}
where $N$ is the number of logarithmic neurons in the preceding layer, and $[\ ]$ denotes the concatenation operation. We then feed $\mathbf{z}_0$ into $L$ hidden layers:
\begin{equation}
\begin{split}
	\mathbf{z}_1 &= ReLU(\mathbf{W}_1\mathbf{z}_0+\mathbf{b}_1) \\
	&~~~~~~~~~~\cdots \cdots \\
	\mathbf{z}_L &= ReLU(\mathbf{W}_L\mathbf{z}_{L-1}+\mathbf{b}_L)
\end{split}
\end{equation}
where $\mathbf{W}_L$ and $\mathbf{b}_L$ denote the weight matrix and bias vector of the $L$-th layer, respectively. $ReLU$ is the Rectifier activation function~\cite{Relu} to capture nonlinear element-wise feature interactions.
At last, the output $\mathbf{z}_L$ of the hidden layers is transformed to the final prediction $\hat{y}$:
\begin{equation}
	\hat{y}=\mathbf{w}_p^\top\mathbf{z}_{L}+b_p
\end{equation}
where $\mathbf{w}_p$ and $b_p$ denote the weight vector and bias term of the prediction layer, respectively.

\subsection{Optimization}

As \model enhances FMs from the perspective of learning adaptive-order cross features, it can be applied to a variety of prediction tasks, including classification, regression and ranking, where the objective functions should be chosen accordingly.  
For binary classification tasks producing 0 or 1 target labels, a common objective function is the logarithmic loss:
\small
\begin{equation}\label{eq:loss}
	Logloss = -\frac{1}{K}\sum_{i=1}^K y_i \log\sigma(\hat{y}_i)+ (1-y_i)\log(1-\sigma(\hat{y}_i))
\end{equation} 
\normalsize
where $K$ is the total number of training instances, and $\sigma$ denotes the sigmoid function. For regression tasks, we can minimize the mean squared error loss. In this work, we focus on binary classification tasks and optimize the log loss in Equation~(\ref{eq:loss}). 
We employ the Adam optimizer~\cite{Adam}, a variant of Stochastic Gradient Descent that dynamically tunes the learning rate during the training process, thus leading to faster convergence~\cite{DBLP:conf/dasfaa/SunSJ0LXW19}.

Besides, we perform batch normalization (BN)~\cite{bn} on the outputs of logarithmic transformation, exponential transformation, and all the hidden layers with two considerations. First, the feature embeddings $\mathbf{e}$ are usually initialized and optimized to be close to zero. After the logarithmic transformation, the embeddings tend to involve large negative values with a significant variance, which is harmful to the optimization of the parameters in the successive layers. As BN can scale and shift the outputs to normalized values, it is crucial to the training process of \modelns. Second, we employ multi-layered neural networks after the logarithmic transformation layer. Performing BN on the outputs of hidden layers helps alleviate the covariance shift problem~\cite{bn}, leading to faster convergence and better model performance empirically.

\subsection{Ensemble \model with DNNs}
Prior works~\cite{wide_and_deep,DeepFM,xDeepFM} have proposed to ensemble the prediction results from cross feature-based models such as FMs with those from raw feature-based neural methods to boost the performance. 
As \model adaptively learns cross features for prediction, we can combine it with deep neural networks (DNNs) in a similar manner. Note that neural structures are powerful to capture nonlinear and deep feature interactions at the element-wise level, which can be viewed as a fine-grained feature interactions modeling method to enhance \modelns.
To enforce the ensemble between \model and the neural networks, we first train the two models separately. After that, we develop an ensemble model to combine the prediction results from the two trained models, as below:
\begin{equation} 
	\hat{y}_{ensem}=w_1\hat{y}_{AFN}+w_2\hat{y}_{DNN}+b
\end{equation}
where $\hat{y}_{AFN}$ and $\hat{y}_{DNN}$ are the predictions made from the trained \model and DNN respectively, $w_1$ and $w_2$ are the corresponding coefficients, and $b$ is a bias term. The ensemble model can be trained by optimizing the log loss similar to Equation (\ref{eq:loss}).
We dub this ensembled model as ``\modelns$^+$''. Our ensemble method is a bit different from the one used by DeepFM~\cite{DeepFM} whose feature embeddings are shared among FM and DNN. Here we separate the embedding layers of \model and DNN to avoid interference. Our primary concern is that unlike DeepFM, the distribution of the embedding values in \modelns, which should always keep positive, is far from that in DNN. The separation slightly increases the model complexity but leads to better performance according to our experiments.

\subsection{Discussions}

\subsubsection{Understanding orders in \modelns.}
\model learns the power (i.e., the orders) of each feature in cross features though the logarithmic transformation layer. As no restriction is enforced on the weight matrix $\mathbf{W}_{LTL}$ of the logarithmic transformation layer, the learned feature orders can be decimals or negative values.
In order to understand the feature orders learned by \modelns, we borrow some ideas from the field-aware factorization machines (FFMs)~\cite{FFM}. In FFMs, each feature is associated with $m$ feature embeddings, where $m$ is the number of feature fields. FFMs is distinct from FMs in that each feature employs different embeddings when interacting with features from different fields. The insight of FFMs is to avoid interference among the feature space of different fields. 
In \modelns, the order of each feature can be considered as a scaling factor for the corresponding feature embedding. For example, consider an embedding with values ranging from $0$ to $1$. An order larger than $1$ would shrink embedding values while an order smaller than $1$ would do the opposite. By analogy to FFMs, the orders learned by \model can be utilized to rescale feature embeddings when interacting with other features in different fields.

\subsubsection{Relation to FMs and HOFMs.}
We first show that FMs can be viewed as a special case of \modelns.
According to Equation (\ref{eq:LTL}), the output $\mathbf{y}_j$ from a logarithmic neuron is able to represent any second-order cross features by setting the power $w_{ij}$ of each feature embedding appropriately.
Suppose we have enough logarithmic neurons to produce all the second-order cross features, and the successive hidden layers approximate a simple summation function at the element-level. Then \model can exactly recover FMs.
Similarly, when we have enough logarithmic neurons to deliver all the cross features within the maximum order and allow the hidden layers to approximate a summation function, \model is able to recover HOFMs. Note that in HOFMs~\cite{hofm}, feature embeddings in different orders can be either shared or learned separately. 

\subsubsection{Time complexity analysis.}
Recall that we use $k$ and $m$ to denote the rank of feature embeddings and the number of feature fields, respectively. 
In \modelns, a logarithmic neuron in Equation (\ref{eq:LTL}) can be computed in ${\rm O}(km)$ time. Assume we use $N$ logarithmic neurons to obtain cross features. The computational complexity of the logarithmic transformation layer is ${\rm O}(kmN)$. Considering the additional cost in the hidden layers, the total time complexity of \model is ${\rm O}(kmN+n_W)$, where $n_W$ is the total number of weights in the hidden layers.
As for HOFMs, supposing $n$ is the maximum order of feature combinations as predefined, it takes ${\rm O}(km^n)$ time to deliver a prediction which can be reduced to
${\rm O}(kmn^2)$ with dynamic programming~\cite{hofm}. 
Note that the time complexity of HOFMs is highly correlated with the maximum order $n$ of cross features, while in \modelns, both $N$ and $n_W$ are only determined by the model structure due to its adaptive cross features generation manner. The time cost of training \model with the optimal setting is empirically close to that of CIN~\cite{xDeepFM}.

\section{Experiments}

In this section, we conduct experiments to answer the following research questions:

\noindent{\bf RQ1:} How do our proposed methods \model and \modelns$^+$ perform against the state-of-the-art methods?

\noindent{\bf RQ2:} How does the performance of \model vary with different settings of the hyper-parameters?

\noindent{\bf RQ3:} What are the learned feature orders in \modelns, and can AFN find useful cross features from data?

\subsection{Experimental Settings}

\subsubsection{Datasets.}
\begin{table}
\small
	\captionsetup{labelfont=bf}
	\centering
		\caption{Statistics of the datasets.}
			\label{tab:datasets}
	\begin{tabular}{cccc}
		\toprule[1pt]
		Dataset&\!\#instances\!&\!\#fields\!&\!\#features\!\\
		\midrule[0.5pt] 
		Criteo&45,840,617&39&2,086,936\\
		Avazu&40,428,967&22&1,544,250\\
		Movielens&2,006,859&3&90,445\\
		Frappe&288,609&10&5,382\\
		\bottomrule[1pt]
	\end{tabular}
\end{table}

\begin{table*}[t!]
\small
	\captionsetup{labelfont=bf}
	\centering
		\caption{Performance comparison.}
		\label{tab:performance}
	\begin{tabular}{llcccccccc}
		\toprule[1pt]
\multirow{2}{*}{Model Class}&
\multirow{2}{*}{Model}
		 &\multicolumn{2}{c}{Criteo}&\multicolumn{2}{c}{Avazu}&\multicolumn{2}{c}{Movielens}&\multicolumn{2}{c}{Frappe}\\
		 &&AUC&Logloss&AUC&Logloss&AUC&Logloss&AUC&Logloss\\
		\midrule[0.5pt] 
		 First-Order&LR&0.7858&0.4636&0.7313&0.4065&0.9215&0.3080&0.9329&0.2860\\
		\midrule[0.5pt] 
		\multirow{2}{*}{Second-Order 
}&
FM& 0.7933&0.4574&0.7496&0.3740&0.9388&0.2797&0.9641&0.2143\\ 
&AFM&0.7953&0.4554&0.7454&0.3766&0.9295&0.2836&0.9639&0.2294\\ 
\midrule[0.5pt]
\multirow{6}{*}{High-Order 
}&
CrossNet&0.7915&0.4585&0.7498&0.3756&0.9323&0.2929&0.9393&0.2835\\ 
&HOFM&0.7960&0.4551&0.7516&0.3756&0.9410&0.3088&0.9709&0.2141\\
&NFM&0.7968&0.4537&0.7531&0.3761& 0.9441&0.3004&0.9727&0.2079\\
&PNN&0.8026&0.4509&0.7526&0.3737&0.9469&0.2792&0.9735&0.2012\\
&CIN&0.8042&0.4472&\bf{0.7533}&0.3756&\bf{0.9494}&\bf{0.2600}&0.9704&0.2342\\ 
&\bf{AFN}&\bf{0.8061}&\bf{0.4458}&0.7512&\bf{0.3731}&0.9477&0.2753&\bf{0.9759}&\bf{0.1784}\\
\midrule[0.5pt]
\multirow{6}{*}{Ensembled 
}&
Deep\&Cross&0.8059&0.4463&0.7550&0.3721&0.9419&0.2791&0.9402&0.2808\\
		&Wide\&Deep&0.8062&0.4453&0.7529&0.3744&0.9381&0.3310&0.9728&0.2038\\
		&DeepFM&0.8025&0.4501&0.7535&0.3742&0.9424&0.3131&0.9719&0.2108\\
		&xDeepFM&0.8070&\bf{0.4443}&0.7535&0.3737&0.9448&0.2717&0.9738&0.2098\\ 
		&\bf{AFN$^+$}&\bf{0.8074}&0.4451&\bf{0.7555}&\bf{0.3718}&\bf{0.9500}&\bf{0.2585}&\bf{0.9783}&\bf{0.1762}\\
		\bottomrule[1pt]
	\end{tabular}  
\end{table*}

\begin{figure*}
	\centering
	\subcaptionbox{Number of logarithmic neurons\label{fig:hyper:a}}
	[.635\columnwidth]{\includegraphics[width=.635\columnwidth]{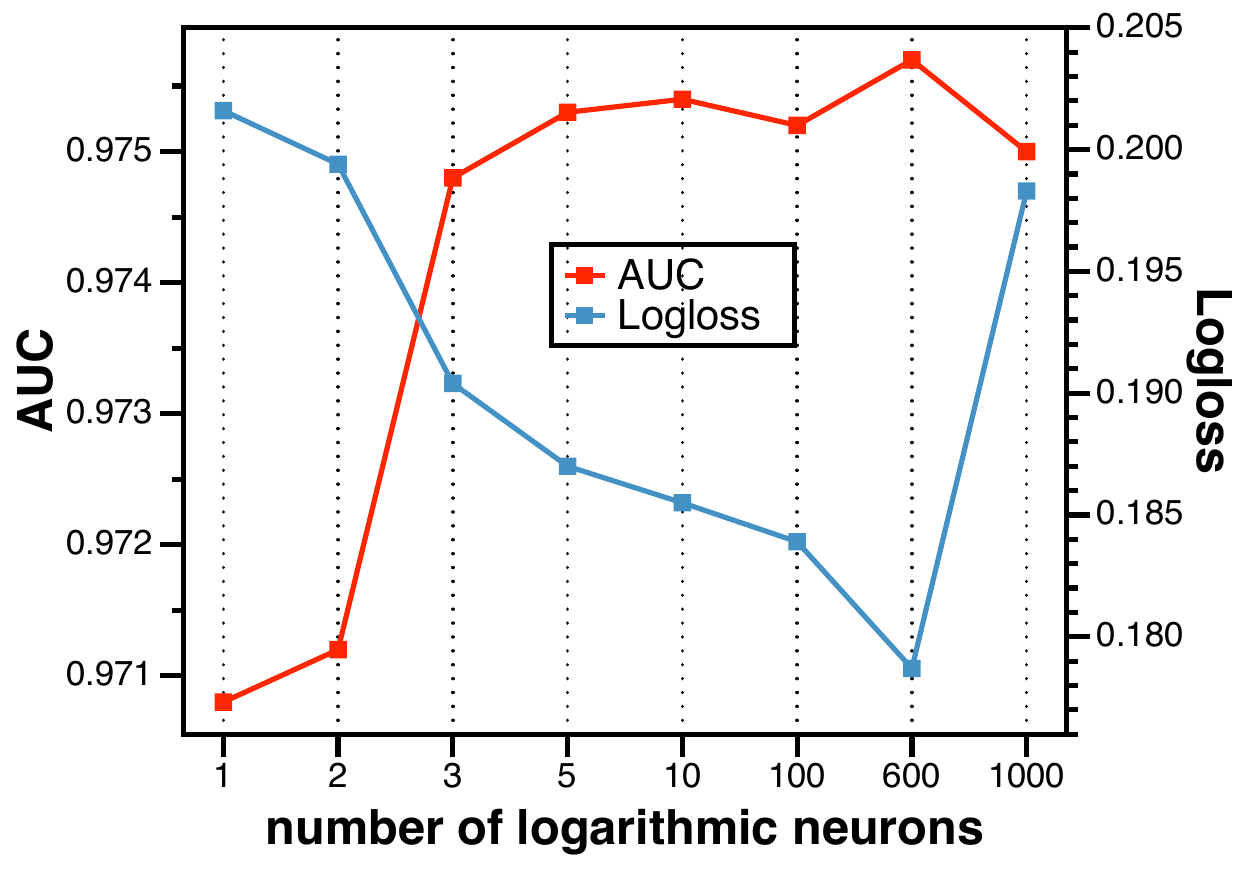}}
	\subcaptionbox{Depth of hidden layers\label{fig:hyper:b}}
	[.635\columnwidth]{\includegraphics[width=.635\columnwidth]{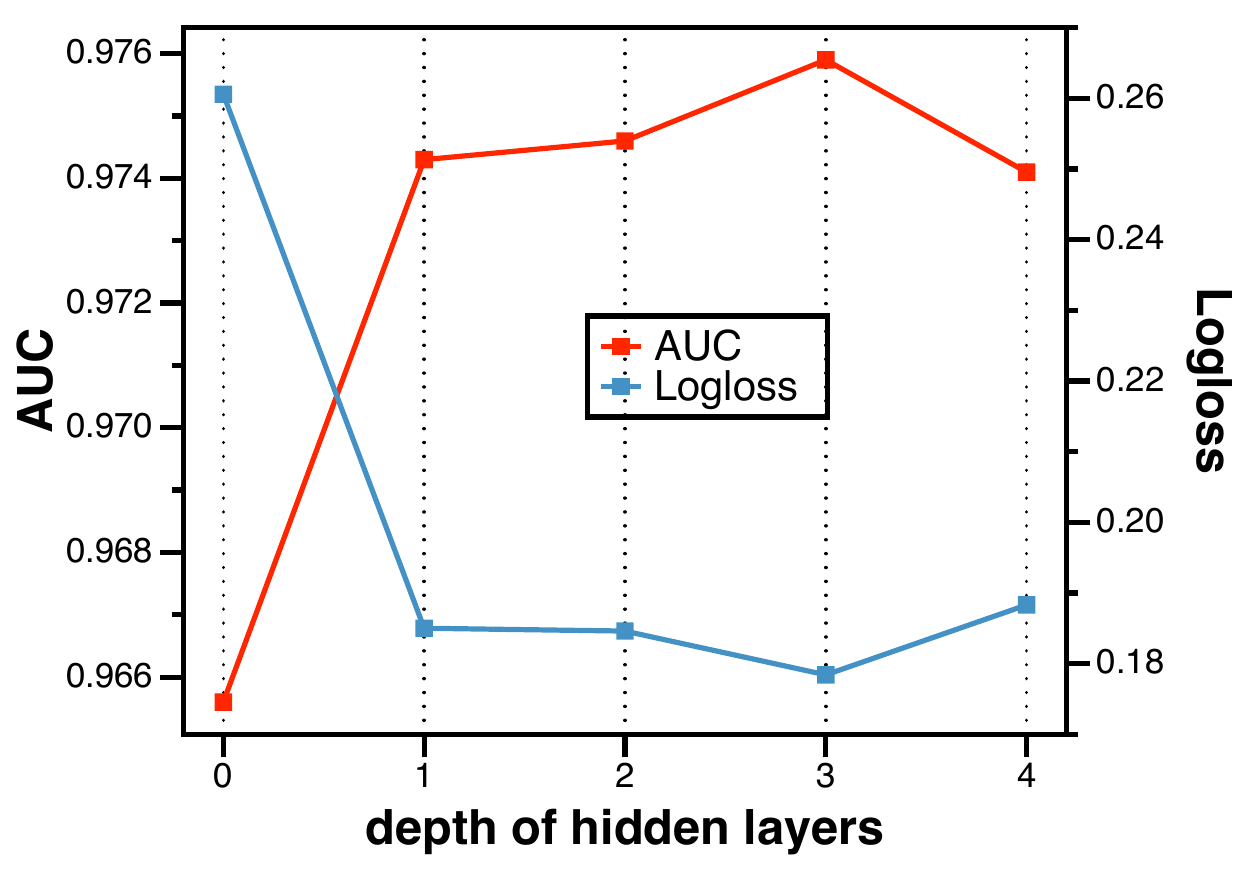}}
	\subcaptionbox{Number of hidden neurons\label{fig:hyper:c}}
	[.635\columnwidth]{\includegraphics[width=.635\columnwidth]{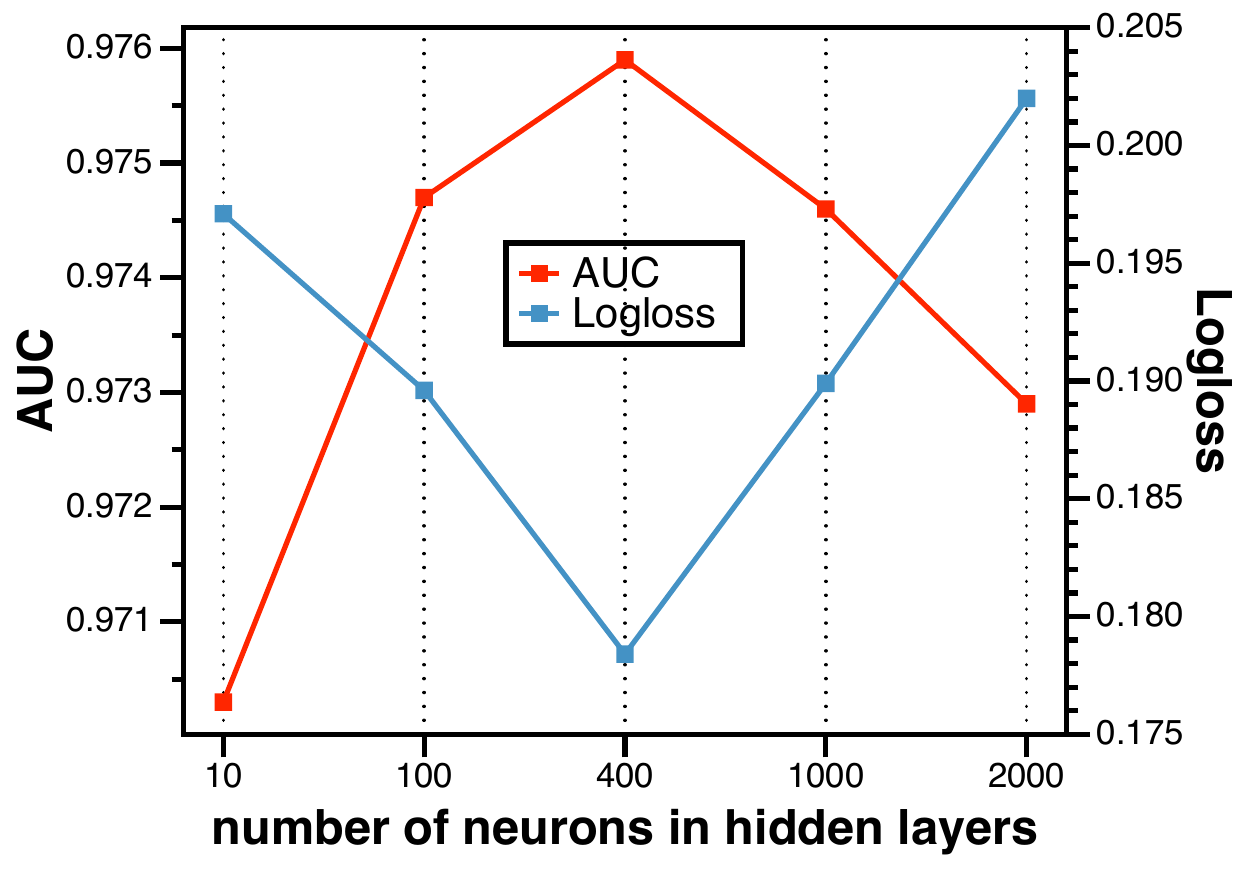}}
	\caption{Effects of hyperparameters on the performance of \modelns.}\label{fig:hyper}
\end{figure*}

We conduct experiments with four publicly accessible datasets following previous works~\cite{xDeepFM,nfm}: Criteo\footnote{\small{\url{http://labs.criteo.com/2014/02/kaggle-display-advertising-challenge-dataset/}}}, Avazu\footnote{\small\url{https://www.kaggle.com/c/avazu-ctr-prediction}}, Movielens\footnote{\small\url{https://grouplens.org/datasets/movielens/}} and Frappe\footnote{\small\url{http://baltrunas.info/research-menu/frappe}}. For each dataset, we randomly split the instances by 8:1:1 for training, validation and test, respectively. The details of the four datasets are summarized in Table~\ref{tab:datasets}.\\
{(1) \bf Criteo}: This is a popular industry benchmarking dataset for CTR prediction, which contains 13 numerical feature fields and 26 categorical feature fields.\\
{(2) \bf Avazu}: This dataset contains users' click records on mobile advertisements. It has 22 feature fields including user features and advertisement attributes.\\
{(3) \bf Movielens}: This dataset consists of users' tagging records on movies. We focus on personalized tag recommendation by converting each tagging record (user ID, movie ID, tag) to a feature vector as input. The target value denotes whether the user has assigned a particular tag to the movie.\\
{(4) \bf Frappe}:  
This dataset contains app usage logs from users under different contexts (e.g., daytime, location). We converted each log (user ID, app ID, context features) to a feature vector as input. The target value indicates whether the user has used the app under the context.

\noindent{\bf{Evaluation metrics.}}
We adopt two metrics for performance evaluation: AUC (Area Under the ROC curve) and Logloss (cross entropy). Note that a slight increase in AUC or decrease in Logloss at .001-level is known to be a significant improvement for the tasks such as CTR prediction~\cite{wide_and_deep,DeepFM,song2018autoint}. 

\noindent{\bf{Comparison methods.}}
We compare \model and \modelns$^+$ with four classes of the existing approaches: (i) first-order approaches that linearly sum up raw features; (ii) FM-based methods that consider second-order cross features; (iii) advanced approaches that model high-order feature interactions; (iv) ensemble models that involve a DNN as the counterpart. We briefly describe these methods as follows. \\
{$\bullet$ \bf Linear Regression (LR)}. It sums up raw features linearly. \\
{$\bullet$ \bf Wide\&Deep}~\cite{wide_and_deep}. It integrates LR with DNN. Note that we omit the hand-crafted cross features for a fair comparison.\\
{$\bullet$ \bf FM}~\cite{libFM}. FM models second-order cross features with factorization techniques for prediction. \\
{$\bullet$ \bf HOFM}~\cite{hofm}. It is the high-order version of FM.\\
{$\bullet$ \bf DeepFM}~\cite{DeepFM}. It is an ensemble between DNN and FM.\\
{$\bullet$ \bf AFM}~\cite{AFM}. It extends FM via the attention mechanism to distinguish the importance of second-order cross features. \\ 
{$\bullet$ \bf CrossNet}~\cite{deep_and_cross}. It explicitly models feature interactions by taking the outer product of input feature vectors.\\
{$\bullet$ \bf Deep\&Cross}~\cite{deep_and_cross}. It is the ensemble between CrossNet and DNN. \\
{$\bullet$ \bf NFM}~\cite{nfm}. It sums up pairwise Hadamard product of input feature vectors followed by fully connected layers. \\
{$\bullet$ \bf PNN}~\cite{pnn}. It models feature interactions by concatenating pairwise inner or outer products of input feature vectors. \\ 
{$\bullet$ \bf CIN}~\cite{xDeepFM}. It produces high-order cross features by computing outer products of feature vectors at different orders.\\
{$\bullet$ \bf xDeepFM}~\cite{xDeepFM}. It integrates CIN with DNN.
 
\subsubsection{Implementation details.}

We implement our methods using Tensorflow\footnote{\small\url{https://github.com/WeiyuCheng/AFN-AAAI-20}}. We apply Adam with a learning rate of 0.001 and a mini-batch size of 4096. The default number of logarithmic neurons is set to 1500, 1200, 800 and 600 for Criteo, Avazu, Movielens and Frappe datasets, respectively. We use 3 hidden layers and 400 neurons per layer by default in \modelns. To avoid overfitting, we perform early-stopping according to the AUC on the validation set. 
We set the rank of feature embeddings to 10 in all the models. We use the same neural network structure (i.e., 3 layers, 400-400-400) for all the approaches that involve DNN for a fair comparison. The maximum order in HOFM is set to 3. All the other hyperparameters are tuned on the validation set. For each empirical result, we run the experiments for 3 times and report the average value. 

\subsection{Performance Comparison (RQ1)}

\noindent{\bf{Comparing with individual models.}}
We first compare \model with various individual models involving first-order, second-order and high-order feature interactions.
The results are shown in Table~\ref{tab:performance}.
We have three important observations. First, \model yields the best or competing performance over all the datasets. For Criteo and Frappe, \model outperforms the second-best model CIN by a large margin, i.e., on average, the increase on AUC and the decrease on log loss are 0.0037 and 0.0286, respectively. For Movielens, \model achieves the second-best performance, and for Avazu, it achieves the best log loss with a moderate AUC. Regarding the good performance of simpler models on Movielens and Avazu, we conjecture that the predictions on these two datasets rely more on lower-order cross features, and the advantages of \model are thus restricted. Note that Movielens only contains three feature fields, and the benefit of finding useful higher-order cross features can be marginal.
Second, \model consistently outperforms FMs and HOFMs on all the datasets, which verifies that learning adaptive-order cross features can bring better predictive performance than modeling fixed-order feature interactions.
Third, the models that utilize higher-order feature interactions generally outperform those based on lower-order cross features, especially when the number of feature fields is large.
This is consistent with the intuition that higher-order feature interactions have stronger predictive power.

\noindent{\bf{Comparing with integrated models.}}
\modelns$^+$ integrates \model and DNN to exploit both explicit cross features and implicit element-wise feature interactions for prediction. We compare \modelns$^+$ with several state-of-the-art ensemble models.
As shown in Table~\ref{tab:performance}, \modelns$^+$ achieves the best performance on the four datasets. On average, \model outperforms xDeepFM, which integrates CIN and DNN, by achieving 0.003 and 0.012 improvements on AUC and log loss, respectively. This demonstrates that the adaptive-order cross features learned by \model are quite different from the implicit feature interactions modeled by DNN, thus improving the performance gain significantly when combining two different types of feature interactions for prediction. 

\begin{figure*}
\centering
\sbox{\tempbox}{

  \includegraphics[width=1.\columnwidth]{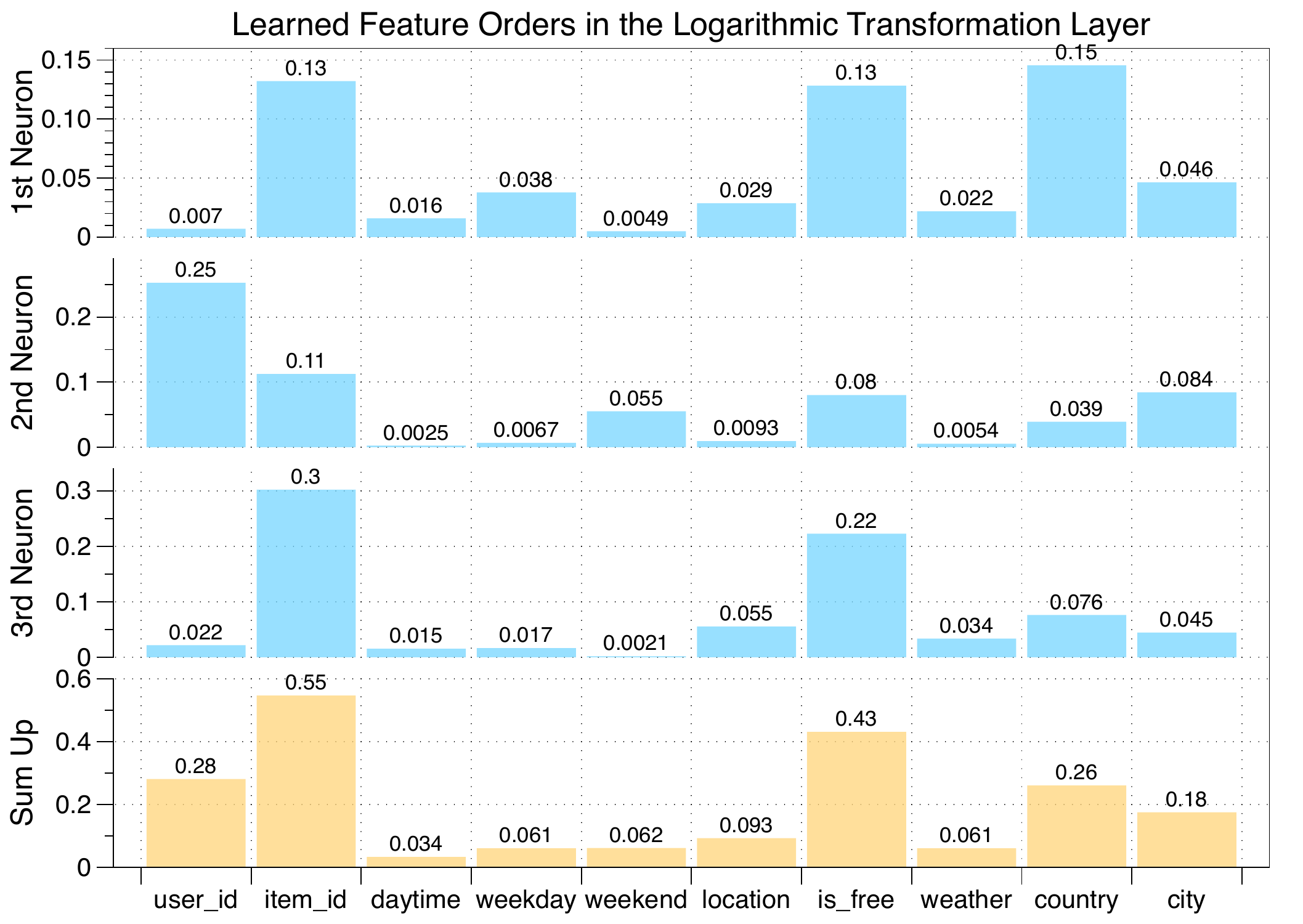}}%

\parbox[b][\ht\tempbox][s]{.6\columnwidth}{%
  \subcaptionbox{Individual feature order\label{fig:weights:a}}{\includegraphics[width=.517\columnwidth]{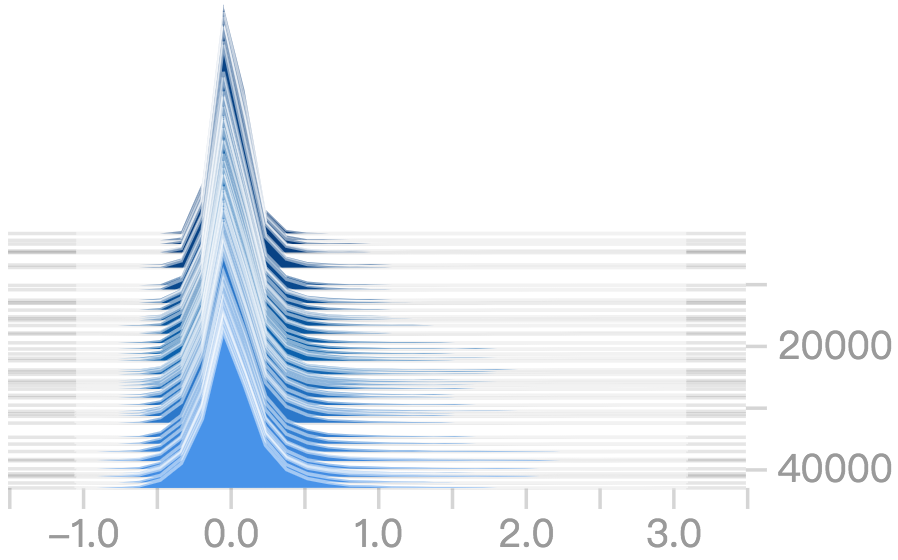}}

  \subcaptionbox{Cross feature order\label{fig:weights:b}}{\includegraphics[width=.517\columnwidth]{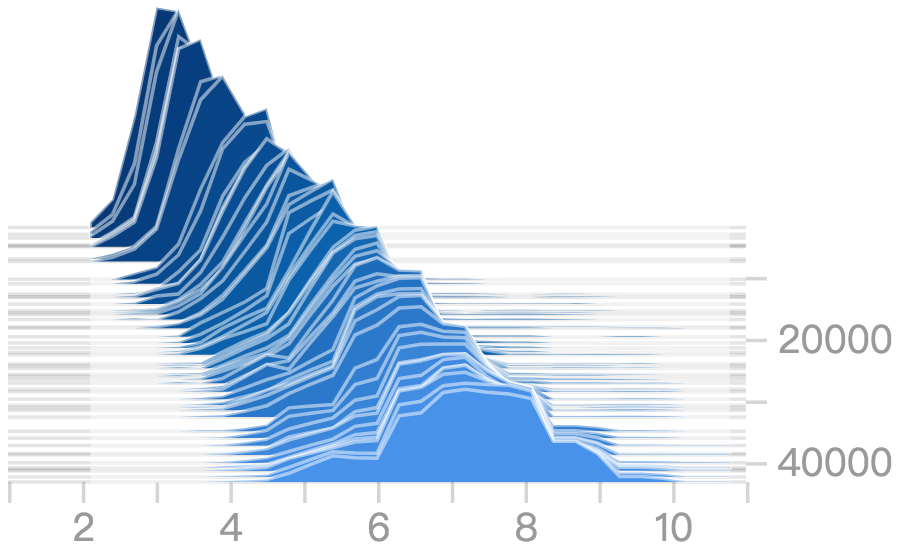}}
}\qquad  
\subcaptionbox{Learned feature orders on each feature field\label{fig:case_study}}{\usebox{\tempbox}}

\caption{Quantitative analysis and case study. (a)\&(b): Distribution of the learned feature orders (x-axis) over training steps (y-axis) on Criteo dataset. (c) A case study on Frappe dataset, where we set the number of logarithmic neurons to $3$.\label{fig:weights}}
\end{figure*}

\subsection{Hyperparameter Investigation (RQ2)}

We now study the effect of three hyperparameters on the performance of \modelns. We only provide the results on Frappe as the results on the other three datasets are similar.

\noindent{\bf{Number of logarithmic neurons}}.
Figure~\ref{fig:hyper:a} provides the results over different numbers of logarithmic neurons in the logarithmic transformation layer. We can see that the performance of \model shows an increasing trend, followed by a decreasing trend when the number of neurons becomes larger. This indicates that an appropriate number of logarithmic neurons should be employed to make a trade-off between expressiveness and generalization to achieve optimal performance. Surprisingly, the advantage of \model is stable even when the number of logarithmic neurons is less than 5. This result demonstrates that finding a small number of discriminative cross features is vital to the prediction accuracy and \model is effective to find these critical cross features.

\noindent{\bf{Depth of hidden layers}}.
Figure~\ref{fig:hyper:b} shows the effect of the depth of hidden layers. We observe that stacking hidden layers upon the learned adaptive-order cross features is beneficial in improving the model performance. However, it is worth noticing that the performance of \model is not highly dependent on the number of hidden layers. When the depth is set to 0, and the prediction is made by a weighted sum over the learned cross features, \model can still achieve fairly good results. This demonstrates the effectiveness of the logarithmic transformation layer in learning discriminative cross features.

\noindent{\bf{Number of neurons in hidden layers}}.
As shown in Figure~\ref{fig:hyper:c}, the performance of \model first grows with the number of neurons. This is because more parameters bring better expressiveness to the model. The performance starts to degrade when the parameter size of the hidden layers exceeds 600, which is caused by overfitting as the training loss keeps on decreasing afterward.

\subsection{Quantitative Analysis and Case Study (RQ3)}

\noindent{\bf{Learned feature orders.}}
We now investigate the learned feature orders in the logarithmic transformation layer of \modelns. Figure~\ref{fig:weights} shows the variation of feature orders during the whole training procedure on Criteo Dataset. From Figure~\ref{fig:weights:a}, we can see that the orders of individual feature field are typically centered around zero and within the range of $[-1,1]$. This is quite different from the typical factorization-based methods where individual feature orders are either $0$ or $1$. The relaxation in the learned feature orders allows the original feature embeddings to be rescaled when composing different cross features. 
We also provide the order distribution of cross features in Figure~\ref{fig:weights:b}, where the order of a cross feature is computed by the sum of absolute values of the constituent features orders. We can see that the learned cross feature orders are gradually optimized during the training process. The final cross feature orders spread over a wide range (from $4$ to $10$), instead of being fixed to a predefined value (e.g., $2$) as in much of the factorization-based work.

\noindent{\bf{Case study.}}
To get a deeper understanding of the cross features learned by \modelns, we conduct a case study on Frappe Dataset, where the description of each feature field is available. For illustration purpose, we limit the number of logarithmic neurons to be $3$. Figure~\ref{fig:case_study} provides the absolute values of individual feature orders on each neuron and the summation. From the figure, we can approximately infer that three cross features \emph{(item\_id, is\_free, country)}, \emph{(user\_id, item\_id)}, and \emph{(item\_id, is\_free)} are learned in the respective logarithmic neurons. Moreover, by summing up feature orders in three neurons, the most discriminative feature fields are found to be \emph{item\_id}, \emph{is\_free}, and \emph{user\_id}. This is reasonable because user and item identities are the most commonly used features in collaborative filtering, and \emph{is\_free}, denoting whether a user has paid for a mobile app, is a strong indicator of users' preferences towards apps.

\section{Conclusion}

In this paper, we introduced the Adaptive Factorization Network (\modelns), which learns arbitrary-order feature interactions adaptively from data. Instead of explicitly modeling all the cross features within a fixed maximum order, \model is able to generate discriminative cross features and the weights of the corresponding features automatically.
The key idea is to transform feature embeddings into a logarithmic space and treat the power of each feature in a feature combination as the coefficient to be learned. 
Feedforward neural networks are further applied to combine the learned cross features for prediction. 
We also showed that \model can generalize FMs and HOFMs with computational efficiency. Extensive experiments on four real-world datasets demonstrate the superior predictive performance of \model compared with the state-of-the-art methods.

\section*{Acknowledgments}
The authors would like to thank the anonymous reviewers for their
insightful reviews.
This work is supported by the National Key Research
and Development Program of China (No. 2018YFC0831604) and NSFC (No. 61602297). Linpeng Huang is also supported by the National Key Research
and Development Program of China (No. 2018YFB1003302).

\bibliographystyle{aaai}
\bibliography{1650-aaai20} 

\end{document}